\documentclass[a4paper]{article}

\usepackage[english]{babel}
\usepackage[latin1]{inputenc}
\usepackage[T1]{fontenc}

\usepackage{yfonts}

\usepackage{mathtools}
\usepackage[pdftex]{graphicx}
\usepackage{epstopdf}
\usepackage{makeidx}
\usepackage{nicefrac}
\usepackage{multicol}
\usepackage[bottom]{footmisc}
\usepackage{comment}
\usepackage[margin=15pt,font=small,labelfont=bf,labelsep=endash,aboveskip=-3pt,belowskip=-8pt]{caption}
\usepackage[most]{tcolorbox}
\usepackage{forest}
\usepackage[e]{esvect}
\usepackage{pdflscape}
\usepackage{pdfpages}
\usepackage{comment}
\usepackage{amsfonts}
\usepackage{amssymb}
\usepackage{amsmath}
\usepackage{amsthm}
\usepackage{algorithm2e}
\usepackage{stmaryrd}

\allowdisplaybreaks

\usepackage{url}
\usepackage{mlmodern}
\usepackage[T1]{fontenc}

\usepackage[small,euler-hat-accent]{eulervm}

\DeclareMathSizes{10}{9.5}{7}{7}

\usepackage{epsfig}
\usepackage{color}

\newcommand {\real}{\mathbb{R}}

\newcommand {\nat}{\mathbb{N}}

\usepackage{bm}
\newcommand{\negmath}[1]{\mathbold{{#1}}}

\newtcolorbox[auto counter, number within=chapter]{example}[1][]{%
	enhanced jigsaw, 
	breakable=true,
	before skip=7pt,
	after skip=7pt,
	left=0mm,
	right=0mm,
	top=1mm,
	bottom=1.5mm,
	boxsep=1.5mm,
	borderline west={1pt}{0pt}{black!23!white}, 
	sharp corners, 
	boxrule=0pt, 
	fonttitle={\small\bfseries},
	coltitle={black},  
	opacityfill=0,
	fontupper=\small,
	title={Example~\thetcbcounter.\ },  
	attach title to upper, 
	 #1
}

\newcounter{th}
\newtcbtheorem[list inside=mteo]{mteo}{Theorem}
{
	enhanced jigsaw,
	arc=0mm,outer arc=.5mm,
	breakable=true,
	colback=black!1!white,
	before skip=5pt,
	after skip=5pt,
	leftrule=0pt,
	rightrule=0pt,
	coltitle=black,
	top=1mm,
	bottom=1.5mm,
	left=0mm,
	right=0mm,
	colframe=black!23!white,
	fonttitle=\bfseries,
	separator sign dash,
	theorem style=standard,
	fontupper=\itshape,
	titlerule=0mm,
	separator sign dash,
	before title={\stepcounter{th}}
}{teo}

\newtcbtheorem[number within=th,list inside=mteo]{mcoro}{Corollary}
{
	enhanced jigsaw,
	arc=0mm,outer arc=.5mm,
	breakable=true,
	colback=black!1!white,
	before skip=5pt,
	after skip=5pt,
	leftrule=0pt,
	rightrule=0pt,
	coltitle=black,
	top=0.5mm,
	bottom=0.5mm,
	left=0mm,
	right=0mm,
	colframe=black!23!white,
	fonttitle=\bfseries,
	separator sign dash,
	theorem style=standard,
	fontupper=\itshape,
	titlerule=0mm,
	separator sign dash
}{cor}

\tcolorboxenvironment{proof}{	
	enhanced jigsaw, 
	breakable=true,
	before skip=5pt,
	after skip=5pt,
	left=1mm,
	right=0mm,
	top=0.5mm,
	bottom=0.5mm,
	boxsep=1mm,
	borderline west={1.5pt}{0pt}{black!23!white}, 
	sharp corners, 
	boxrule=0pt, 
	fonttitle={\small\bfseries},
	coltitle={black},  
	opacityfill=0,
	attach title to upper, 
}

\newcommand {\map}[3]{#1 : #2 \mapsto #3}

\newcommand {\vto}[1]{\negmath{#1}}

\newcommand {\vun}[1]{\negmath{\hat{#1}}}

\newcommand {\vtf}[1]{\pmb{#1}}

\newcommand {\mat}[1]{\mathit{#1}}

\newcommand {\ete}[2]{\bar{\tnr{\mathcal{#1}}}^{#2}}
\newcommand {\etel}[3]{\bar{\tnr{\mathcal{#1}}}^{#2}_{(#3)}}

\newcommand {\tnr}[1]{\vtf{#1}}

\title{High-Order Tensor Regression in Sparse Convolutional Neural Networks}

\author{Roberto Dias Algarte\thanks{robertodias70@outlook.com}}

\begin{document}

\RestyleAlgo{ruled}

\maketitle

\begin{abstract}
\noindent 
This article presents a generic approach to convolution that significantly differs from conventional methodologies in the current Machine Learning literature. The approach, in its mathematical aspects, proved to be clear and concise, particularly when high-order tensors are involved. In this context, a rational theory of regression in neural networks is developed, as a framework for a generic view of sparse convolutional neural networks, the primary focus of this study. As a direct outcome, the classic Backpropagation Algorithm is redefined to align with this rational tensor-based approach and presented in its simplest, most generic form.
\end{abstract}

\section{Convolution Over High-Order Tensors}
In this study, we shall work only with real Hilbert spaces, which are complete normed inner product vector spaces defined by real scalars where the inner product induces the norm. Since every $n$-dimensional Hilbert space $\tnr{\mathcal{U}}^{n}$ has orthonormal basis $\{\vun{u}_1,\cdots,\vun{u}_n\}$, an arbitrary vector $\vto{u}\in \tnr{\mathcal{U}}^{n}$ is decomposed as 
\begin{equation}
	\vto{u}=\sum_{i=1}^n u_i\vun{u}_i\,,
\end{equation}
where scalars $u_i$ are the coordinates of $\vto{u}$ on the basis under consideration. Given the collection of Hilbert spaces $\tnr{\mathcal{U}}_1^{n_1},\tnr{\mathcal{U}}_2^{n_2},\cdots,\tnr{\mathcal{U}}_q^{n_q}$, the real vector space  $\ete{U}{q}:=\tnr{\mathcal{U}}_1^{n_1}\otimes\cdots\otimes\tnr{\mathcal{U}}_q^{n_q}$ is here called a \emph{tensor space} of order $q$, whose elements are called \emph{tensors}. For the practical purposes of our study, a tensor is considered to be a magnitude of generic order, in the sense that it is called a vector if its order is one, that is, $\bar{\tnr{\mathcal{U}}}^1$ is a simple Hilbert space. By definition, scalars are tensors of order zero. Note that a tensor space can be defined by the same $n$-dimensional Hilbert space, that is, if there are $q$ repetitions of $\tnr{\mathcal{U}}^{n}$ in $\tnr{\mathcal{U}}^{n}\otimes\cdots\otimes\tnr{\mathcal{U}}^{n}$, then it is obvious that this tensor space also has order $q$ and each of its constituent vector spaces has the same dimension, which $\ete{U}{q}$, as previously defined, may not necessarily have.

An arbitrary tensor of $\ete{U}{q}$ is constituted by vectors $\vto{u}^i$ in each of its constituent vector spaces $\tnr{\mathcal{U}}_i^{n_i}$, being represented by $\vto{u}^1\otimes\cdots\otimes\vto{u}^q$. Since a tensor space is itself a vector space, this tensor is decomposed as 
\begin{equation}
	\vto{u}^{1}\otimes\cdots\otimes\vto{u}^q=\sum_{i_1=1}^{n_1}\cdots\sum_{i_q=1}^{n_q} (\vto{u}^1\otimes\cdots\otimes\vto{u}^q)_{i_1i_2\cdots i_q}\vun{u}_{i_1}^1\otimes\cdots\otimes\vun{u}_{i_q}^q\,,
\end{equation}
where the scalars are the coordinates of the tensor and $\{\vun{u}_1^i,\cdots,\vun{u}_{n_i}^i\}$ is an orthonormal basis of the $i$-th order of $\ete{U}{q}$. In the particular case where most of its coordinates are zero, the tensor is said to be \emph{sparse}. In our tensor representation, the practical goal of operation $\otimes$, called \emph{tensor product}, is to compose two magnitudes of order one (vector) in order to generate a magnitude of order two (second order tensor). Now, let $q=3$ in order to present the following properties.
\begin{itemize}
	\item [i.]  Multiplication of a tensor by a scalar $\alpha\in\real$: 
	\begin{equation}
		\alpha(\vto{u}^1\otimes\vto{u}^2\otimes\vto{u}^3) = (\alpha\vto{u}^1)\otimes\vto{u}^2\otimes\vto{u}^3= \vto{u}^1\otimes(\alpha\vto{u}^2)\otimes\vto{u}^3= \vto{u}^1\otimes\vto{u}^2\otimes(\alpha\vto{u}^3)\,;
	\end{equation} 
	\item [ii.] Distributivity of the tensor product on vector sum:  
		\begin{equation}
		\vto{u}^1\otimes\vto{u}^2\otimes(\vto{u}^3_1+\vto{u}^3_2) =
		\vto{u}^1\otimes\vto{u}^2\otimes\vto{u}^3_1+\vto{u}^1\otimes\vto{u}^2\otimes\vto{u}^3_2 \,.
	\end{equation} 
\end{itemize}
A basis of our tensor space $\ete{U}{q}$ is defined to be the set of tensors
\begin{equation}
	\{\vun{u}_{1}^1\otimes\cdots\otimes\vun{u}_{1}^q,\vun{u}_{1}^1\otimes\cdots\otimes\vun{u}_{2}^q,\cdots,\vun{u}_{1}^1\otimes\cdots\otimes\vun{u}_{n_q}^q,\cdots,\vun{u}_{n_1}^1\otimes\cdots\otimes\vun{u}_{n_q}^q\}\,,
\end{equation}  
where the indices run sequentially from right to left along the basis tensors. The dimension of a tensor space is the defined to be product of the dimensions of its constituent vector spaces, that is,
\begin{equation}
\dim(\ete{U}{q}) = \prod_{i=1}^{q} \dim(\tnr{\mathcal{U}}_i^{n_i})\,.  
\end{equation}   
Sometimes, in order to simplify notation, we represent a tensor with a single uppercase bold letter, that is, if arbitrary tensor $\tnr{T}=\vto{u}^1\otimes\cdots\otimes\vto{u}^q$ then $\tnr{T}\in\ete{U}{q}$. 

Given a tensor space $\bar{\tnr{\mathcal{K}}}^{r}:=\tnr{\mathcal{K}}_1^{m_1}\otimes\cdots\otimes\tnr{\mathcal{K}}_r^{m_r}$, we consider the set
\begin{equation}
\ete{U}{q}\otimes\bar{\tnr{\mathcal{K}}}^{r}: = \tnr{\mathcal{U}}_1^{n_1}\otimes\cdots\otimes\tnr{\mathcal{U}}_q^{n_q}\otimes\tnr{\mathcal{K}}_1^{m_1}\otimes\cdots\otimes\tnr{\mathcal{K}}_r^{m_r}  
\end{equation}  
a tensor space of order $q+r$. In this context, given a tensor space  $\bar{\tnr{\mathcal{K}}}^{r}\otimes\ete{V}{m}$, the $r$-\emph{order inner product} of a tensor 
\begin{equation}
\tnr{T}=\vto{u}^1\otimes\cdots\otimes\vto{u}^q\otimes\vto{x}^1\otimes\cdots\otimes\vto{x}^r\in\ete{U}{q}\otimes\bar{\tnr{\mathcal{K}}}^{r}
\end{equation}
and a tensor
\begin{equation}
	\tnr{S}=\vto{y}^1\otimes\cdots\otimes\vto{y}^r\otimes\vto{v}^1\otimes\cdots\otimes\vto{v}^m\in\bar{\tnr{\mathcal{K}}}^{r}\otimes\ete{V}{m}
\end{equation}
is a tensor of order $q+m$
\begin{equation}\label{eq:genInner}
\tnr{T}\odot_r\tnr{S}:=(\prod_{i=1}^r\vto{x}^i\cdot\vto{y}^i)\,\vto{u}^1\otimes\cdots\otimes\vto{u}^q\otimes\vto{v}^1\otimes\cdots\otimes\vto{v}^m\,.
\end{equation}
Note that the $r$-order inner product is not always commutative; moreover, tensor spaces $\ete{U}{q}$ and $\ete{V}{m}$ need not necessarily be defined. From the definition above, the following properties can be straightforwardly obtained:
\begin{itemize}
	\item [i.] $\vto{x}_1\odot_1\vto{y}_1=\vto{x}_1\cdot\vto{y}_1$;
\item [ii.] $(\vto{u}\otimes\vto{x}_1)\odot_1\vto{y}_1=\vto{y}_1\odot_1(\vto{x}_1\otimes\vto{u})=(\vto{x}_1\cdot\vto{y}_1)\vto{u}$;
	\item [iii.] Given $\tnr{K}_1,\tnr{K}_2\in\bar{\tnr{\mathcal{K}}}^{m}$, then $\tnr{K}_1\odot_r\tnr{K}_2$ exists only if $r\leqslant m$;
	\item[iv.] Given $\tnr{K}_1,\tnr{K}_2\in\bar{\tnr{\mathcal{K}}}^{r}$, then  $\tnr{K}_1\odot_r\tnr{K}_2=\tnr{K}_2\odot_r\tnr{K}_1$ is a scalar. 
\end{itemize}
Considering the last property, since the $r$-order inner product behaves like an ordinary inner product of vectors, we usually define $\tnr{K}_1\cdot\tnr{K}_2=\tnr{K}_1\odot_r\tnr{K}_2$. Now, if a tensor space $\ete{U}{q}\otimes\ete{U}{q}$ is considered, there is a special element $\tnr{I}\in\ete{U}{q}\otimes\ete{U}{q}$, called \emph{identity tensor}, where 
\begin{equation}
	\tnr{I}\odot_q\tnr{T}=\tnr{T}\odot_q\tnr{I}=\tnr{T}\,,\, \forall\, \tnr{T}\in\ete{U}{q}\,.
\end{equation}
Note that, from this definition, only tensor spaces of even order have an identity tensor. Sometimes, it is useful to identify the order of the identity tensor by adopting the notation $\tnr{I}^{2q}$. By tedious algebraic manipulations, it is possible to arrive at
\begin{equation}\label{eq:identity}
	\tnr{I}^{2q} = \sum_{i_1=1}^{n_1} \cdots\sum_{i_q=1}^{n_q} \vun{u}_{i_1}^{1}\otimes\cdots\otimes\vun{u}_{i_q}^{q}\otimes\vun{u}_{i_1}^{1}\otimes\cdots\otimes\vun{u}_{i_q}^{q}\,.
\end{equation}

It is also useful for our study to represent a vector $\vto{u}$ and a second order tensor $\vto{u}\otimes\vto{v}$ as \emph{arrays} of order $1$ and $2$ the following way:
\begin{equation}
	[\vto{u}]:=\begin{bmatrix}
		u_1 & \cdots & u_n\\
	\end{bmatrix}^T\,;
\end{equation}
\begin{equation}
	[\vto{u}\otimes\vto{v}]:=
	\begin{bmatrix}
		(\vto{u}\otimes\vto{v})_{11} & \cdots & (\vto{u}\otimes\vto{v})_{1m}\\
		\vdots & \ddots & \vdots\\
		(\vto{u}\otimes\vto{v})_{n1} & \cdots & (\vto{u}\otimes\vto{v})_{nm}
	\end{bmatrix}\,.
\end{equation}
Generically, given a tensor $\tnr{T}=\vto{u}^{1}\otimes\cdots\otimes\vto{u}^q$ and its coordinates $T_{i_1\cdots i_q}$, the elements of array $[\tnr{T}]$ of order $q$ are defined by 
\begin{equation}
	[\tnr{T}]_{i_1\cdots i_q}=T_{i_1\cdots i_q}\,.
\end{equation}
Sometimes it is useful to say that the \emph{size} of array $[\tnr{T}]$ is $n_1\times\cdots\times n_q$, where value $n_i$ is here called the $i$-th \emph{dimension} of $[\tnr{T}]$. Since coordinates of tensors are involved, these array representations above are obviously basis dependent. Moreover, given  $0<k_i\leqslant h_i$ where $h_i<n_i$, the array 
\begin{equation}
	[\tnr{T}]_{k_1:h_1,\cdots,k_q:h_q}
\end{equation}
is called a \emph{sub-array} of $[\tnr{T}]$ that is constituted by contiguous coordinates of $\tnr{T}$ from indexes $k_i$ to $h_i$, inclusive, in each order of the tensor. In other words, this sub-array has size $(h_1-k_1+1)\times\cdots\times (h_q-k_q+1)$ and 
\begin{equation}
	[\tnr{T}]_{k_1:h_1,\cdots,k_q:h_q} = [\sum_{i_1=k_1}^{h_1}\cdots\sum_{i_q=k_q}^{h_q}T_{i_1\cdots i_q}\vun{u}^{1}_{i_1-k_1+1}\otimes\cdots\otimes\vun{u}^{q}_{i_q-k_q+1}]\,.
\end{equation}

In our study, \emph{convolution} is an operation defined between an array and a tensor of the same order. In order to grasp its fundamental concepts, we present them in the convolution of a one order array over a vector, which is the simplest form of convolution.  Considering $\mat{F}$ an array of order one, called \emph{kernel} or \emph{filter}, with dimension $k>0$ and a vector $\vto{u}\in\tnr{\mathcal{U}}^{n}$, where $n>k$, the convolution of $\mat{F}$ over $\vto{u}$ is the vector $\mat{F}\varoast\vto{u}\in\tnr{\mathcal{U}}^{\bar{n}}$,  $\bar{n}:=n-k+1$,  defined by
\begin{equation}\label{eq:conv1D}
	\mat{F}\varoast\vto{u} = F(\mat{F})\odot_1\vto{u}
\end{equation}
in which $F$ is a tensor valued linear function, called \emph{compounded filter function}, with the rule
\begin{equation}\label{eq:compFunc}
	F(\mat{X}) = \sum_{i=0}^{\bar{n}-1}\sum_{j=1}^{k} \mat{X}_j \vun{u}_{1+i}\otimes \vun{u}_{j+i}\,,
\end{equation}
that maps an array of order one to a tensor of $\tnr{\mathcal{U}}^{\bar{n}}\otimes\tnr{\mathcal{U}}^{n}$, where $\{\vun{u}_1,\cdots,\vun{u}_{n}\}$ is an orthonormal basis of $\tnr{\mathcal{U}}^{n}$. As a simple example, for $k=2$ and $n=5$, it is straightforward to obtain that 
\begin{align}
	\mat{F}\varoast\vto{u} =& (\mat{F}_{1}u_1+\mat{F}_{2}u_2)\vun{u}_1+(\mat{F}_{1}u_2+\mat{F}_{2}u_3)\vun{u}_2+\nonumber\\
	&+(\mat{F}_{1}u_3+\mat{F}_{2}u_4)\vun{u}_3+(\mat{F}_{1}u_4+\mat{F}_{2}u_5)\vun{u}_4\,,
\end{align}
from which the intuitive concept of convolution can be clearly observed: the coordinates of the convolution are built by sliding the filter elements one position rightward along the coordinates of $\vto{u}$. In this context of one position sliding, we say that the \emph{stride} $s$ of the convolution is one. The previous equality written in matrix representations is 
\begin{equation}
	[\mat{F}\varoast\vto{u}] = [F(\mat{F})]\cdot [\vto{u}] = \begin{bmatrix}
		\mat{F}_1 & \mat{F}_2 & 0 & 0 & 0\\
		 0 & \mat{F}_1 & \mat{F}_2 & 0 & 0\\
		0 & 0 & \mat{F}_1 & \mat{F}_2 & 0\\
		0 & 0 & 0 & \mat{F}_1 & \mat{F}_2\\
	\end{bmatrix}\cdot\begin{bmatrix}
	u_1\\
	u_2\\
	u_3\\
	u_4\\
	u_5
	\end{bmatrix}\,,
\end{equation}
where $[F(\mat{F})]$ is clearly a sparse Toeplitz matrix. The following properties are straightforwardly obtained from \eqref{eq:conv1D} and \eqref{eq:compFunc}: 
\begin{itemize}
	\item [i.] $\mat{F}\varoast(\vto{x}+\vto{y})=\mat{F}\varoast\vto{x}+\mat{F}\varoast\vto{y}$;
	\item [ii.] $(\mat{F}_1+\mat{F}_2)\varoast\vto{x}=\mat{F}_1\varoast\vto{x}+\mat{F}_2\varoast\vto{x}$;
	\item [iii.] if $k=1$ and $s=1$, $\mat{F}\varoast\vto{x}=\mat{F}_1\vto{x}\,$.
\end{itemize}
Considering a generic stride, that is, $0<s\leqslant n-k$, the compounded filter function   \eqref{eq:compFunc} must be adjusted to 
\begin{equation}
	F(\mat{X}) = \sum_{i=0}^{\bar{n}-1}\sum_{j=1}^{k} \mat{X}_j \vun{u}_{1+s\cdot i}\otimes \vun{u}_{j+s\cdot i}\,,
\end{equation}
where 
\begin{equation}
\bar{n}:= \left\lfloor \dfrac{n-k}{s}\right\rfloor + 1\,.
\end{equation}

As shown so far, it is important to observe that convolution of filters with dimensions greater than one, which is the most usual form of convolution, is an operation that decreases or \emph{down-samples} the dimension of the tensor involved, preserving its order. In the current one order case, a vector of dimension $n$ is down-sampled to a vector of dimension  $\bar{n} < n$ through filter $\mat{F}$. Nevertheless, there is a sagacious strategy, called \emph{zero padding}, to obtain from vector $\vto{u}$ a convolution result of dimension $n$ by defining a vector $\vto{u}^*$ with dimension $n^*\geqslant n$. In this sense, considering $n^*:=(n-1)s+k$, let a vector $\vto{u}^*\in\tnr{\mathcal{U}}^{n^*}$ be defined by
\begin{equation}
\vto{u}^*=\sum_{i=1}^{n}u_{i}\vun{u}^*_{i+g}\,,
\end{equation}
where 
\begin{equation}
	g=\left\lceil\dfrac{n^*-n}{2}\right\rceil
\end{equation}
and the set of orthonormal vectors  $\{\vun{u}^*_1,\cdots,\vun{u}^*_{n^*}\}$ is a basis of $\tnr{\mathcal{U}}^{n^*}$. In this context, $\dim(\mat{F}\varoast\vto{u}^*)=\dim(\vto{u})$ or
\begin{equation}
	\left\lfloor \dfrac{n^*-k}{s}\right\rfloor + 1 = n\,.
\end{equation}
In zero padding, definition \eqref{eq:conv1D} becomes   
\begin{equation}
	\mat{F}\varoast\vto{u}^* = F(\mat{F})\odot_1\vto{u}^*
\end{equation}
and if stride $s=1$, the first filter element $\emph{F}_1$ slides over all coordinates of $\vto{u}$, which does not occur when there is no zero padding (\emph{valid padding}).

Using the same intuitive concept presented in the previous convolution over a vector, we now extend it to the convolution of a filter $\mat{F}$, with size $k_1\times k_2$, over a second order tensor $\tnr{T}\in\tnr{\mathcal{U}}^{n_1}\otimes\tnr{\mathcal{U}}^{n_2}$. This operation is then defined by the expression   
\begin{equation}\label{eq:conv2D}
	\mat{F}\varoast\tnr{T} = F(\mat{F})\odot_2\tnr{T}\,,
\end{equation}
where 
\begin{equation}
	F(\mat{X}) = \sum_{i_1=0}^{\bar{n}_1-1}\sum_{i_2=0}^{\bar{n}_2-1}\sum_{j_1=1}^{k_1}\sum_{j_2=1}^{k_2} \mat{X}_{j_1j_2} \vun{u}_{1+s_1\cdot i_1}^1\otimes \vun{u}_{1+s_2\cdot i_2}^2\otimes\vun{u}_{j_1+s_1\cdot i_1}^1\otimes \vun{u}_{j_2+s_2\cdot i_2}^2 \,.
\end{equation}
In this definition, $0<k_i\leqslant n_i$,   $0<s_i\leqslant n_i-k_i$ and 
\begin{equation}
	\bar{n}_i:= \left\lfloor \dfrac{n_i-k_i}{s_i}\right\rfloor + 1\,.
\end{equation}
Note that the stride here is constituted by two values instead of one because the filter can slide in two different directions over $\tnr{T}$. For the case of zero padding, $n_i^*=(n_i-1)s_i+k_i$, $g_i=\lceil(n_i^*-n_i)/2\rceil$ and 
\begin{equation}
	\tnr{T}^*=\sum_{i_1=1}^{n_1}\sum_{i_2=1}^{n_2}T_{i_1i_2}\vun{u}^{*1}_{i_1+g_1}\otimes\vun{u}^{*2}_{i_2+g_2}\,.
\end{equation}

Still considering the conditions above for a generic order, if the convolution of an array $\mat{F}$ of size $k_1\times\cdots\times k_r$ is performed over a tensor $\tnr{T}\in\tnr{\mathcal{U}}^{n_1}\otimes\cdots\otimes\tnr{\mathcal{U}}^{n_r}$, this operation is defined by 
\begin{equation}\label{eq:convRD}
	\mat{F}\varoast\tnr{T} = F(\mat{F})\odot_r\tnr{T}\,,
\end{equation}
where 
\begin{align}\label{eq:compounded}
	F(\mat{X}) =& \sum_{i_1=0}^{\bar{n}_1-1}\cdots\sum_{i_r=0}^{\bar{n}_r-1}\sum_{j_1=1}^{k_1}\cdots\sum_{j_r=1}^{k_r} \mat{X}_{j_1\cdots j_r} \vun{u}_{1+s_1\cdot i_1}^1\otimes\cdots\otimes \nonumber\\ &\otimes\vun{u}_{1+s_r\cdot i_r}^r\otimes\vun{u}_{j_1+s_1\cdot i_1}^1\otimes\cdots\otimes \vun{u}_{j_r+s_r\cdot i_r}^r \,.
\end{align}
For the case of zero padding, tensor $\tnr{T}$ defines 
\begin{equation}
	\tnr{T}^*=\sum_{i_1=1}^{n_1}\cdots\sum_{i_r=1}^{n_r}T_{i_1\cdots i_r}\vun{u}^{*1}_{i_1+g_1}\otimes\cdots\otimes\vun{u}^{*r}_{i_r+g_r}\,.
\end{equation}

\section{Sparse Neural Network Regression}

Considering the mathematical apparatus presented in the previous section, let $\ete{U}{q}$ and $\ete{V}{q}$ be tensor spaces. In the context of the tensor mapping $\map{\vtf{f}}{\ete{U}{q}}{\ete{V}{q}}$, we specify that the set of ordered pairs
\begin{equation}\label{eq:train}
	\pmb{\mathcal{T}}=\{ (\vto{T}_{[1]},\vtf{f}(\vto{T}_{[1]})),\cdots, (\vto{T}_{[p]},\vtf{f}(\vto{T}_{[p]}))\}\,,\, p\in\nat^*\,,
\end{equation}
called a training set, is completely known. In simple terms, Supervised Machine Learning (SML) is basically a set of techniques to find an approximate function $\widetilde{\vtf{f}}\approx\vtf{f}$ through a training set $\pmb{\mathcal{T}}$. Since the specified co-domain $\ete{V}{m}$ is a real vector space, that is, since it is constituted by ``continuous'' magnitudes, we call the SML problem under study a regression problem. Following a similar approach presented in \cite{algarte_2024} for an approximation mapping $\map{\widetilde{\vtf{f}}}{\ete{U}{q}}{\ete{V}{q}}$, we study a deep neural network of $k>3$ layers by defining 
\begin{equation}
	\widetilde{\vtf{f}} = \widetilde{\vtf{f}}^{(k)}\circ\cdots\circ\widetilde{\vtf{f}}^{(1)}\circ\widetilde{\vtf{f}}^{(0)}\,,
\end{equation}    
where $\widetilde{\vtf{f}}^{(0)}$ is the identity tensor function, $\map{\widetilde{\vtf{f}}^{(1)}}{\ete{U}{q}}{\etel{Z}{q}{1}}$ and 
$\map{\widetilde{\vtf{f}}^{(k)}}{\etel{Z}{q}{k-1}}{\ete{V}{q}}$. Given $0<l\leqslant k$, the notation $\etel{Z}{q}{l}$ refers to a $q$-th order tensor space on which layer $l$ is defined and we specify that $\etel{Z}{q}{k}=\ete{V}{q}$. The Artificial Neural Network (ANN) strategy specifies mappings 
\begin{equation*}
\map{\widetilde{\vtf{f}}^{(l)}}{\etel{Z}{q}{l-1}}{\etel{Z}{q}{l}}\,,\, 0< l \leqslant k\,,
\end{equation*}
whose functions $\widetilde{\vtf{f}}^{(l)}$, called \emph{$l$-th constituent functions} of $\widetilde{\vtf{f}}$,  are described by 
\begin{equation}\label{eq:intermm}
	\widetilde{\vtf{f}}^{(l)}(\tnr{X})=\vtf{\phi}^{(l)}(\tnr{W}^{(l)}\odot_{q}\tnr{X}+\tnr{B}^{(l)})\,.
\end{equation}
where $\tnr{W}^{(l)}\in\etel{Z}{q}{l}\otimes\etel{Z}{q}{l-1}$ is a weight tensor and $\tnr{B}^{(l)}\in\etel{Z}{q}{l}$ is a bias tensor. The function $\vtf{\phi}^{(l)}$ is a differentiable operator, called \emph{activation function}, defined in tensor mapping $\map{\vtf{\phi}^{(l)}}{\etel{Z}{q}{l}}{\etel{Z}{q}{l}}$, with rule
 \begin{equation}\label{eq:activat}
	\vtf{\phi}^{(l)}(\tnr{X}) = \sum_{i_1=1}^{n_1^{(l)}}\cdots\sum_{i_{q}=1}^{n_{q}^{(l)}} \phi^{(l)}(X_{i_1i_2\cdots i_{q_l}})\,\vun{z}^{(l),1}_{i_1}\otimes\cdots\otimes\vun{z}_{i_{q}}^{(l),q}\,,
\end{equation}
where the $j$-th order unitary vectors $\vun{z}_{i_{j}}^{(l),j}$ constitute bases tensors of $\etel{Z}{q}{l}$. The scalar function $\phi^{(l)}$ refers to a typical activation function for regression problems like the Sigmoid, ReLU, Softmax and others. In our study, the weight tensor $\tnr{W}^{(l)}$ is partially known because it is specified to be sparse. So, the neural networks under study here are not dense, that is, every unit of layer $l$ depends on a few specific units of layer $l-1$. Since sparse arrays can be stored efficiently and require fewer calculations, this approach drastically reduces the consumption of computational resources, which is a critical necessity for handling high volume training data. However, this strategy only makes sense in problems where most of the possible unit to unit dependencies can be disregarded, such as image pixel processing. 

Considering the bivariate differentiable mapping  $\map{\tnr{\psi}^{(k)}}{\ete{V}{q}\otimes\etel{Z}{q}{k-1}\times\ete{V}{q}}{\real}$ and the training set \eqref{eq:train}, ANN models define criteria for accepting approximation functions according to the rule   
\begin{equation}\label{eq:batchLoss}
	\tnr{\psi}^{(k)}(\tnr{X},\tnr{Y})=\dfrac{1}{p}\sum_{i=1}^{p}\tnr{\mathcal{L}}[\underbrace{\vtf{\phi}^{(k)}(\tnr{X}\odot_q\tnr{Z}^{(k-1)}_{[i]}+\tnr{Y})}_{\widetilde{\vtf{f}}(\tnr{T}_{[i]})},\vtf{f}(\tnr{T}_{[i]})]\,,
\end{equation}
where 
\begin{equation}\label{eq:layerValue}
\tnr{Z}_{[i]}^{(k-1)}=\widetilde{\vtf{f}}^{(k-1)}\circ\cdots\circ\widetilde{\vtf{f}}^{(0)}(\vto{T}_{[i]})
\end{equation}
and $\tnr{\psi}^{(k)}$ is called a batch loss function, whose arguments are the unknowns of the problem, and $\tnr{\mathcal{L}}$ a scalar-valued differentiable function called simply loss function. Considering scalar $m=\dim(\ete{V}{q})$, we present below typical examples of loss functions $\tnr{\mathcal{L}}$ for tensors.
\begin{itemize}
	\item [i.] Mean Squared Error: 
	\begin{equation}
		MSE(\tnr{X},\tnr{Y})=\dfrac{1}{m}\|\tnr{X}-\tnr{Y}\|^2\,;
	\end{equation}
	\item [ii.] Mean Absolute Error: 
	\begin{equation}
		MAE(\tnr{X},\tnr{Y})=\dfrac{1}{m} \sum_{i_1=1}^{n_1}\cdots\sum_{i_{q}=1}^{n_{q}}  |X_{i_1i_2\cdots i_{q}}-Y_{i_1i_2\cdots i_{q}}|\,;
	\end{equation}
	\item [iv.] Log-Cosh: 
	\begin{equation}
		LCH(\tnr{X},\tnr{Y})=\dfrac{1}{m}\sum_{i_1=1}^{n_1}\cdots\sum_{i_{q}=1}^{n_{q}}\log(\cosh(X_{i_1i_2\cdots i_{q}}-Y_{i_1i_2\cdots i_{q}})\,;
	\end{equation}
	\item [v.] Mean Squared Logarithmic Error: 
	\begin{equation}
		MSLE(\tnr{X},\tnr{Y})=\dfrac{1}{m}\sum_{i_1=1}^{n_1}\cdots\sum_{i_{q}=1}^{n_{q}}[\log(X_{i_1i_2\cdots i_{q}}+1)-\log(Y_{i_1i_2\cdots i_{q}}+1)]^2\,;
	\end{equation}
	\item [vi.] Poisson: 
	\begin{equation}
		POI(\tnr{X},\tnr{Y})=\dfrac{1}{m}\sum_{i_1=1}^{n_1}\cdots\sum_{i_{q}=1}^{n_{q}} X_{i_1i_2\cdots i_{q}}-X_{i_1i_2\cdots i_{q}}\log(Y_{i_1i_2\cdots i_{q}})\,.
	\end{equation}
\end{itemize}

\subsection{Convolutional Neural Networks} 

A sparse layer ANN architecture is called a Convolutional Neural Network (CNN) when the systematic selection of unit to unit dependencies in weight tensors is made via convolution. Considering the previous development, in a CNN, we consider a filter $\mat{F}^{(l)}$ an unknown since the weight tensor 
\begin{equation}
	\tnr{W}^{(l)} := F^{(l)}(\mat{F}^{(l)})\,,
\end{equation}
where $F^{(l)}$ is a differentiable compounded filter function that maps an array of order $q$ to a tensor of $\etel{Z}{q}{l}\otimes\etel{Z}{q}{l-1}$ according to rule \eqref{eq:compounded}. In other words, 
\begin{align}\label{eq:cnnFilter}
	F^{(l)}(\mat{X}) =& \sum_{i_1=0}^{\bar{n}_1^{(l)}-1}\cdots\sum_{i_q=0}^{\bar{n}_q^{(l)}-1}\sum_{j_1=1}^{k_1^{(l)}}\cdots\sum_{j_q=1}^{k_q^{(l)}} \mat{X}_{j_1\cdots j_q} \vun{z}^{(l),1}_{1+s_1\cdot i_1}\otimes\cdots\otimes \nonumber\\ &\otimes\vun{z}^{(l),q}_{1+s_q\cdot i_q}\otimes\vun{z}^{(l-1),1}_{j_1+s_1\cdot i_1}\otimes\cdots\otimes \vun{z}^{(l-1),q}_{j_q+s_q\cdot i_q} \,,
\end{align}
where $0<k_i^{(l)}\leqslant n_i^{(l)}$,   $0<s_i^{(l)}\leqslant n_i^{(l)}-k_i^{(l)}$ and 
\begin{equation}
	\bar{n}_i^{(l)}= \left\lfloor \dfrac{n_i^{(l)}-k_i^{(l)}}{s_i^{(l)}}\right\rfloor + 1\,.
\end{equation}
In CNNs, rule \eqref{eq:intermm} of the $l$-th constituent function becomes 
\begin{equation}\label{eq:approxConv}
	\widetilde{\vtf{f}}^{(l)}(\tnr{X})=\vtf{\phi}^{(l)}(F^{(l)}(\mat{F}^{(l)})\odot_{q}\tnr{X}+\tnr{B}^{(l)})
\end{equation}  
or, in other words,  
\begin{equation}
	\widetilde{\vtf{f}}^{(l)}(\tnr{X})=\vtf{\phi}^{(l)}(\mat{F}^{(l)}\varoast{\tnr{X}}+\tnr{B}^{(l)})\,,
\end{equation}  
where filter $\mat{F}^{(l)}$ and bias $\tnr{B}^{(l)}$ are unknowns\footnote{In our tensor-based approach of convolution and CNN, concepts like local receptive field, weight sharing, feature consistency and others are superfluous or senseless.}. From these equalities, it is important to recall that, in the context of zero padding, the transformed units 
\begin{equation}
	\tnr{X}^{*}=\sum_{i_1=1}^{\bar{n}_1^{(l-1)}}\cdots\sum_{i_q=1}^{\bar{n}_q^{(l-1)}}X_{i_1\cdots i_q}\vun{z}^{*(l-1),1}_{i_1+g_1}\otimes\cdots\otimes\vun{z}^{*(l-1),q}_{i_q+g_q}
\end{equation}
enables the first element of filter $\mat{F}^{(l)}$ to slide over all units $\tnr{X}\in\etel{Z}{q}{l-1}$. 

In order to increase the representational capacity of the layer convolution, particularly the input layer, and also to reduce information bottlenecks, a strategy of convolving multiple filters in a layer is usually adopted. Considering $m>1$ filters, we define the mapping $\map{\vtf{g}_\alpha^{(l)}}{\etel{Z}{q}{l-1}}{\etel{Z}{q}{l}}$ where $\vtf{g}_\alpha^{(l)}$ is called the \emph{feature map} of filter $\alpha$ and function rule is     
\begin{equation}
\vtf{g}_\alpha^{(l)}(\tnr{X})=\mat{F}_\alpha^{(l)}\varoast\tnr{X}+\tnr{B}_\alpha^{(l)}\,,\, \alpha=1,\cdots,m\,.
\end{equation}
The values of $\vtf{g}_\alpha^{(l)}$ are usually called \emph{pre-activation} tensors because they belong to the activation function domain. From the above $m$ expressions, 
we can define a mapping $\map{\vtf{\varphi}^{(l)}}{\etel{Z}{q+1}{l}}{\etel{Z}{q+1}{l}}$, where 
\begin{equation}
\tnr{\varphi}^{(l)}(\tnr{X}) = \sum_{\alpha=1}^m\vtf{\phi}^{(l)}\circ\vtf{g}_\alpha^{(l)}(\tnr{X})\otimes \vun{u}_\alpha\,
\end{equation}
from which values $[\widetilde{\vtf{f}}^{(l)}(\tnr{X})]_\alpha=\vtf{\phi}^{(l)}\circ\vtf{g}_\alpha^{(l)}(\tnr{X})$ are obtained. Therefore, the image of $\tnr{\varphi}^{(l)}$ constitutes the domain for the next layer operations. Since filters $\mat{F}_\alpha^{(l)}$ are unknown magnitudes, an adequate representational capacity is guaranteed by adopting different initializations for each filter $\alpha$; during training, distinct gradient signals further specialize the filters, yielding diverse feature detectors. In order to simplify future mathematical expressions, we shall keep working with a single feature map per layer.

In certain CNN regression problems where data resolution is not critical, it is convenient to perform  down-sampling of certain layers, either to further reduce computational load or to avoid overfitting. A common strategy to down-sample a layer is to decrease its dimension while preserving its order, a strategy called \emph{pooling}: let the operator in $\map{\tnr{\rho}^{(l)}}{\etel{Z}{q}{l}}{\etel{Z}{q}{l}}$ be described by
\begin{equation}\label{eq:pooling}
\tnr{\rho}^{(l)}(\tnr{X}) = \sum_{i_1=1}^{\tilde{n}_1^{(l)}}\cdots\sum_{i_q=1}^{\tilde{n}_q^{(l)}}\rho\left(X_{i_1:i_1+\tilde{k}_1^{(l)},\cdots,i_q:i_q+\tilde{k}_q^{(l)}}\right)\vun{z}^{(l),1}_{i_1}\otimes\cdots\otimes\vun{z}^{(l),q}_{i_q}\,,
\end{equation}
where $0<\tilde{k}_i^{(l)}\leqslant n_i^{(l)}$, $\tilde{n}_i^{(l)}=n_i^{(l)}-\tilde{k}_i^{(l)} + 1$
and $\rho$ is a scalar valued function whose argument is a sub-array of $[\tnr{X}]$. Thereby, given a certain layer $l-1$ to be down-sampled by pooling, rule \eqref{eq:approxConv} becomes
\begin{equation}
	\widetilde{\vtf{f}}^{({l})}(\tnr{X})=\vtf{\phi}^{({l})}(\mat{F}^{({l})}\varoast\tnr{\rho}^{({l}-1)}(\vto{X}))+\tnr{B}^{({l})})\,.
\end{equation}  
When it is not applied in all hidden layers, pooling is usually applied in the first or in the last hidden layers. Moreover, zero padding can also be used in pooling. The most common scalar valued functions $\rho$ for rule \eqref{eq:pooling} are presented below.
\begin{itemize}
	\item [i.] Max pooling: 
\begin{equation}
	\rho({\mat{X}}) = \max(\mat{X});
\end{equation}

	\item [ii.] Average pooling: if ${k}_1\times\cdots\times {k}_q$ is the size of array $\mat{X}$, 
\begin{equation}
	\rho({\mat{X}}) = \dfrac{1}{\prod_{i=1}^q{k}_i}\sum_{i_1=1}^{{k}_1}\cdots\sum_{i_q=1}^{{k}_q} \mat{X}_{i_1\cdots i_q}\,.
\end{equation}
\end{itemize}

In most of the classical ANN regression problems, the number of units in a layer is a specified quantity. From the very nature of the convolutional approach in CNNs, this number is dependent on other specifications, namely, the dimension $\prod_{i=1}^qn_i$ of the input tensor, the sizes $k_1\times\cdots\times k_q$ of the filters, the strides $(s_1,\cdots,s_q)$ of the convolution, the presence of padding and pooling.

\paragraph{The Backpropagation Algorithm.}
In this section, the Backpropagation algorithmic scheme presented in \cite{algarte_2024} for ANNs of low order tensors is here expanded to high order tensors and adapted to CNNs. The algorithm proceeds in iterations, each one constituted by a forward pass, when unit values are calculated forwardly through the layers, followed by a backward pass, when filters and biases of the next iteration are calculated backwardly. In each CNN layer, the first forward pass of the algorithm guesses zero or small values to biases and adopts to filters strategies like Random Guess, Xavier-Glorot, Kaiming, etc. At the last layer, considering an arbitrary pair $(\vto{T}_{[i]},\vtf{f}(\vto{T}_{[i]}))$ of the training set \eqref{eq:train}, the approximated value of $\vtf{f}(\vto{T}_{[i]})$ is 
\begin{equation}\label{eq:approxF}
	\widetilde{\vtf{f}}(\vto{T}_{[i]})=\vtf{\phi}^{(k)}(\tnr{W}^{(k)}\odot_{q}[\widetilde{\vtf{f}}^{(k-1)}\circ\cdots\circ\widetilde{\vtf{f}}^{(0)}(\vto{T}_{[i]})]+\tnr{B}^{(k)})\,.
\end{equation}
The backward pass of the algorithm requires the calculation of the gradient of the loss function $\tnr{\mathcal{L}}$, defined by  \eqref{eq:batchLoss}, from the training set. Thereby, considering 
\begin{equation}
\tnr{Z}_{[i]}^{(l-1)}=\widetilde{\vtf{f}}^{(l-1)}\circ\cdots\circ\widetilde{\vtf{f}}^{(0)}(\vto{T}_{[i]})\,,
\end{equation}
let variable $\tnr{K}^{(l)}_{[i]}:=\tnr{W}^{(l)}\odot_q\tnr{Z}_{[i]}^{(l-1)}+\tnr{Y}$. Then, let $\map{\tnr{\mathcal{\hat{L}}}}{\etel{Z}{q}{k-1}\times\ete{V}{q}}{\real}$ be a mapping with function rule
\begin{equation}\label{eq:newLoss}
\tnr{\mathcal{\hat{L}}}(\tnr{X})=\tnr{\mathcal{L}}[\tnr{X},\vtf{f}(\tnr{T}_{[i]})]\,,
\end{equation}
where variable $\tnr{X}:=\vtf{\phi}^{(k)}(\tnr{K}^{(k)}_{[i]})$. From classical Tensor Calculus, 
\begin{equation}\label{eq:gradiente}
	\underbrace{\nabla(\tnr{\mathcal{\hat{L}}}\circ\vtf{\phi}^{(k)})}_{\tnr{\delta}^{(k)}}(\tnr{K}^{(k)}_{[i]})=\nabla\vtf{\phi}^{(k)}(\tnr{K}^{(k)}_{[i]})\odot_q\nabla\tnr{\mathcal{\hat{L}}}\circ\vtf{\phi}^{(k)}(\tnr{K}^{(k)}_{[i]})\,,
\end{equation}
where function $\tnr{\delta}^{(k)}\in\ete{V}{q}$ measures the contribution of layer $k$ to loss value $\tnr{\mathcal{L}}[\widetilde{ \vtf{f}}(\tnr{T}_{[i]}),\vtf{f}(\tnr{T}_{[i]})]$. Considering the hidden layers $r=k-1,\cdots,1$ of the backward pass, we specify
\begin{equation}\label{eq:backSoul}
	\nabla\tnr{\mathcal{\hat{L}}}\circ\vtf{\phi}^{(r)}(\tnr{K}^{(r)}_{[i]})=\tnr{\delta}^{(r+1)}(\tnr{K}^{(r+1)})\odot_q\tnr{W}^{(r+1)}\,.
\end{equation}
Considering expression \eqref{eq:gradiente} for layer $r$, we can define that 
\begin{equation}
	\tnr{\delta}^{(r)}(\tnr{K}^{(r)}_{[i]})=\nabla\vtf{\phi}^{(r)}(\tnr{K}^{(r)}_{[i]})\odot_q (\tnr{\delta}^{(r+1)}(\tnr{K}^{(r+1)}_{[i]})\odot_q\tnr{W}^{(r+1)}\,,
\end{equation}
where the weight tensor $\tnr{W}^{(r+1)}$ is known from the forward pass. Definition \eqref{eq:backSoul} is the \emph{raison d'\^etre}  of the Backpropagation Algorithm since it propagates the contribution $\tnr{\delta}^{(k)}$ backwardly to all hidden layers. From the batch loss  rule \eqref{eq:batchLoss} adapted to the CNN context and rule \eqref{eq:newLoss}, we can write that
\begin{equation}
	\tnr{\psi}^{(r)}(\mat{X},\tnr{Y})=\dfrac{1}{p}\sum_{i=1}^{p}\tnr{\mathcal{\hat{L}}}\circ\vtf{\phi}^{(r)}(F^{(r)}(\mat{X})\odot_q\tnr{Z}^{(r-1)}_{[i]}+\tnr{Y})\,,
\end{equation}
where filter $\mat{X}$ and bias $\tnr{Y}$ are the unknowns of the CNN problem. Considering expression \eqref{eq:gradiente} for layer $r$, the following partial gradients of $\tnr{\psi}^{(r)}$ can be obtained from the chain and product rules of Tensor Calculus:
\begin{align}\label{eq:gradX}
	\nabla_{\mat{X}}\tnr{\psi}^{(r)}&=\dfrac{1}{p}\sum_{i=1}^{p}\nabla_{\mat{X}}(F^{(r)}(\mat{X})\odot_q\tnr{Z}^{(r-1)}_{[i]}+\tnr{Y})\odot_q\nabla\vtf{\phi}^{(r)}(\tnr{K}^{(r)}_{[i]})\odot_q\nabla\tnr{\mathcal{\hat{L}}}\circ\vtf{\phi}^{(r)}(\tnr{K}^{(r)}_{[i]})\nonumber\\
	&=\dfrac{1}{p}\sum_{i=1}^{p}\tnr{Z}^{(r-1)}_{[i]}\odot_q\nabla F^{(r)}(\mat{X})\odot_q\tnr{\delta}^{(r)}(\tnr{K}_{[i]}^{(r)})
\end{align}
and
\begin{equation}\label{eq:gradY}
	\nabla_{\tnr{Y}}\tnr{\psi}^{(r)}=\dfrac{1}{p}\sum_{i=1}^{p}\tnr{\delta}^{(r)}(\tnr{K}_{[i]}^{(r)})
\end{equation}
where definitions $\nabla_{\bullet}\tnr{\psi}^{(r)}:=\nabla_{\bullet}\tnr{\psi}^{(r)}(\mat{X},\tnr{Y})$ result in constant functions because the coordinates of the gradient $\nabla F^{(r)}(\mat{X})\in\etel{Z}{q}{r-1}\otimes\etel{Z}{q}{r}\otimes\etel{Z}{q}{r}$ are
\begin{equation}
[\nabla F^{(r)}(\mat{X})]_{t_1\cdots t_qi_1\cdots i_qj_1\cdots j_q } =  \dfrac{\partial [F^{(r)}(\mat{X})]_{i_1\cdots i_qj_1\cdots j_q}}{\partial \mat{X}_{t_1\cdots t_q} }=(\delta_{i_1t_1}\cdots\delta_{i_qt_q})_{j_1\cdots j_q}\,,
\end{equation}
where $\delta$ is the Kronecker Delta. This previous equality leads to the constant tensor $\nabla F^{(r)}:=\nabla F^{(r)}(\mat{X})$ described\footnote{This result is obtained from the contraction property of the Kronecker Delta.} by
\begin{align}
	\nabla F^{(r)} =& \sum_{i_1=0}^{\bar{n}_1^{(r)}-1}\cdots\sum_{i_q=0}^{\bar{n}_q^{(r)}-1}\sum_{j_1=1}^{k_1^{(r)}}\cdots\sum_{j_q=1}^{k_q^{(r)}} \vun{z}^{(r-1),1}_{j_1+s_1\cdot i_1}\otimes\cdots\otimes \vun{z}^{(r-1),q}_{j_q+s_q\cdot i_q}\otimes\nonumber\\
	&\otimes\vun{z}^{(r),1}_{1+s_1\cdot i_1}\otimes\cdots\otimes\vun{z}^{(r),q}_{1+s_q\cdot i_q}\otimes \vun{z}^{(r),1}_{j_1+s_1\cdot i_1}\otimes\cdots\otimes \vun{z}^{(r),q}_{j_q+s_q\cdot i_q}\,.
\end{align}

For the next iteration $t+1$ of the algorithm, it is now possible to obtain new filter $\mat{F}_{t+1}^{(r)}$ and bias $\tnr{B}_{t+1}^{(r)}$ from their values on the current iteration $t$ and the constant gradients \eqref{eq:gradX} and \eqref{eq:gradY}. In other words, 
\begin{alignat}{3} 
	\mat{F}^{(r)}_{t+1}=\tnr{\omega}_{\mat{F}}(\nabla_{\mat{X}}\tnr{\psi}^{(r)},\mat{F}^{(r)}_t) &\quad \text{and}\quad &  \tnr{B}^{(r)}_{t+1}=\tnr{\omega}_{\tnr{B}}(\nabla_{\tnr{Y}}\tnr{\psi}^{(r)},\tnr{B}^{(r)}_t)\,,
\end{alignat}
where functions $\tnr{\omega}_{\mat{F}}$ and $\tnr{\omega}_{\tnr{B}}$ are called optimizers. The simplest example of an optimizer for regression problems is the gradient descent:
\begin{alignat}{3} 
	\mat{F}^{(r)}_{t+1}= \mat{F}^{(r)}_t + \gamma\nabla_{\mat{X}}\tnr{\psi}^{(r)} &\quad \text{and}\quad &  \tnr{B}^{(r)}_{t+1}=\tnr{B}^{(r)}_t + \gamma \nabla_{\tnr{Y}}\tnr{\psi}^{(r)}\,,
\end{alignat}
where $\gamma$ is the learning rate. There are other, more efficient, examples of optimizers, like RMSProp, Adam, Nadam, etc.

In a batch mode of processing, it is convenient to gather up training data tensors $\tnr{T}_{[i]}$ into a single $(q+1)$-th order tensor
\begin{equation}
\tnr{\hat{Z}}^{(0)} := \sum_{i_1=1}^{n_1}\cdots\sum_{i_q=1}^{n_q}\sum_{j=1}^{p}[\tnr{T}_{[j]}]_{i_1\cdots i_q}\vun{u}^{1}_{i_1}\otimes\cdots\otimes\vun{u}^{q}_{i_q}\otimes\vun{u}^{q+1}_{j}\,.
\end{equation}
In this context, the bias tensors $\tnr{B}^{(l)}$ must be changed to $(q+1)$-th order tensors
\begin{equation}
	\tnr{\hat{B}}^{(l)} = \sum_{i_1=1}^{\bar{n}_1^{(l)}}\cdots\sum_{i_q=1}^{\bar{n}_q^{(l)}}\sum_{j=1}^{p}[\tnr{B}^{(l)}]_{i_1\cdots i_q}\vun{u}^{1}_{i_1}\otimes\cdots\otimes\vun{u}^{q}_{i_q}\otimes\vun{u}^{q+1}_{j}\,,
\end{equation}
in which the coordinates $\tnr{B}^{(l)}_{i_1\cdots i_q}$ are repeated in every dimension of the order $j$.  Thereby, adapting $\eqref{eq:approxF}$ to both of these previous tensors, we have    
\begin{equation}
	\tnr{\hat{Z}}^{(l)} =\vtf{\hat{\phi}}^{(l)}(F^{(l)}(\mat{F}^{(l)})\odot_{q}{\tnr{\hat{Z}}^{(l-1)}}+\tnr{\hat{B}}^{(l)})\,,
\end{equation}
where 
\begin{equation}
	\vtf{\hat{\phi}}^{(l)}(\tnr{X}) = \sum_{i_1=1}^{\bar{n}_1^{(l)}}\cdots\sum_{i_{q}=1}^{\bar{n}_{q}^{(l)}}\sum_{j=1}^{p} \phi^{(l)}(X_{i_1i_2\cdots i_{q_l}j})\,\vun{z}^{(l),1}_{i_1}\otimes\cdots\otimes\vun{z}_{i_{q}}^{(l),q}\otimes\vun{z}_{j}^{(l),q+1}\,
\end{equation}
and, as a consequence, coordinates
\begin{equation}\label{eq:zValues}
	\tnr{\hat{Z}}^{(l)}_{i_1\cdots i_qj} = (\tnr{\hat{Z}}^{(l)}_{[j]})_{i_1\cdots i_q}\,.
\end{equation}
Once $\tnr{\hat{Z}}^{(k)}$ is calculated and a loss function is specified, from \eqref{eq:gradiente}, it is possible to build tensor 
\begin{equation}
	\tnr{\hat{\delta}}^{(k)} := \sum_{i_1=1}^{\bar{n}_1^{(k)}}\cdots\sum_{i_{q}=1}^{\bar{n}_{q}^{(k)}}\sum_{j=1}^{p} [\tnr{\delta}^{(k)}(\tnr{K}_{[j]}^{(r)})]
	_{i_1i_2\cdots i_{q_l}}\,\vun{z}^{(k),1}_{i_1}\otimes\cdots\otimes\vun{z}_{i_{q}}^{(k),q}\otimes\vun{z}_{j}^{(k),q+1}\,,
\end{equation}
from which all tensors $\tnr{\hat{\delta}}^{(r)}$ of the backward pass can be calculated. Thereby, from equality $\eqref{eq:zValues}$, the gradients $\nabla_{\mat{X}}\tnr{\psi}^{(r)}$ and $\nabla_{\tnr{Y}}\tnr{\psi}^{(r)}$ are obtained and then, through the specified optimizers, new iteration values for filters and biases are calculated. Now, given the validation set 
\begin{equation}\label{eq:vali}
	\bar{\pmb{\mathcal{T}}}=\{ (\bar{\tnr{T}}_{[1]},\vtf{f}(\bar{\tnr{T}}_{[1]})),\cdots, (\bar{\tnr{T}}_{[\bar{p}]},\vtf{f}(\bar{\tnr{T}}_{[\bar{p}]}))\}\,,\, \bar{p}\in\nat^*,
\end{equation}
at the end of the forward pass in which an epoch $e$ is entirely processed, we calculate the batch loss value
\begin{equation}
	\vartheta_e:=\dfrac{1}{\bar{p}}\sum_{i=1}^{\bar{p}}\tnr{\mathcal{L}}[\widetilde{\vtf{f}}(\bar{\tnr{T}}_{[i]}),\vtf{f}(\bar{\tnr{T}}_{[i]})]\,.
\end{equation}
In this context, the most common stopping criterion for the Backpropagation Algorithm specifies that it ends when either a maximum number of epochs is reached or when a tolerance $\epsilon$ is achieved, that is,    
\begin{equation}\label{eq:tolerance}
	| \vartheta_e - \vartheta_{e-1} | \leqslant \epsilon \,,
\end{equation}
where $\epsilon$ is a small positive value. Note that in a batch mode of processing, every iteration corresponds to an epoch. Algorithm 1 describes this whole procedure step by step.

\begin{algorithm}
	\SetNoFillComment
	\caption{CNN Backpropagation Algorithm in Batch Mode}
	\KwData{Training set $\pmb{\mathcal{T}}$}
	\KwData{Validation set $\bar{\pmb{\mathcal{T}}}$}
	\KwResult{$\{(\mat{F}^{(1)},\tnr{\hat{B}}^{(1)}),\cdots,(\mat{F}^{(k)},\tnr{\hat{B}}^{(k)}))\}$}
	Specify $k$, $\{{\phi}^{(1)},\cdots,{\phi}^{(k)}\}$, $\tnr{\mathcal{L}}$, $\tnr{\omega}$\;
	Specify $\epsilon$ and the maximum number $\alpha$ of epochs\; 
	Initialize  $\{(\mat{F}^{(1)}_1,\tnr{\hat{B}}^{(1)}_1),\cdots,(\mat{F}^{(k)}_1,\tnr{\hat{B}}^{(k)}_1))\}$\;
	Build $\tnr{Z}^{(0)}$\; 
	$c\gets MAX\_FLOAT$\;
	$\vartheta_0\gets MAX\_FLOAT$\;
	$t\gets 1$\;
	\While{$c > \epsilon$ and $t\leqslant \alpha$}{
		$l\gets 1$\;
		\tcc{Forward pass}
		\While{$l\leqslant k$}{
			$\tnr{\hat{Z}}^{(l)}_t =\vtf{\hat{\phi}}^{(l)}(F^{(l)}(\mat{F}^{(l)}_t)\odot_{q}{\tnr{\hat{Z}}^{(l-1)}_t}+\tnr{\hat{B}}^{(l)}_t)$\;
			$l\gets l+1$\;
		}
		Calculate $\vartheta_t $ from $\bar{\pmb{\mathcal{T}}}$ and $\tnr{\mathcal{L}}$\;
		$c\gets |\vartheta_t-\vartheta_{t-1}|$\;
		\If{$c \geqslant \epsilon$}{
			Calculate $\tnr{\hat{\delta}}^{(k)}_t$ from $\tnr{\hat{Z}}^{(k)}_t$ and $\tnr{\mathcal{L}}$\;
			$r\gets k-1$\;
			\tcc{Backward pass}
			\While{$r > 0$}{
				Calculate $\tnr{\hat{\delta}}^{(r)}_t$ from $\tnr{\hat{\delta}}^{(r+1)}_t$ and $\mat{F}^{(r+1)}_t$\;
				Calculate $[\nabla_{\tnr{X}}\tnr{\psi}^{(r)}]_t$ and $[\nabla_{\vto{y}}\tnr{\psi}^{(r)}]_t$ from $\tnr{\hat{\delta}}^{(r)}_t$\;
				Calculate $\mat{F}^{(r)}_{t+1}$ and $\tnr{\hat{B}}^{(r)}_{t+1}$ from $\tnr{\omega}$\;
				$r\gets r -1$\;
			}
		}
		$t\gets t+1$\;
	}
\end{algorithm}

\end{document}